%% file: main.tex
\documentclass[10pt,twocolumn,letterpaper]{article}

 \usepackage[pagenumbers]{cvpr} 

\input{preamble}
\definecolor{cvprblue}{rgb}{0.21,0.49,0.74}
\usepackage[pagebackref,breaklinks,colorlinks,allcolors=cvprblue]{hyperref}

\usepackage{hyperref}
\usepackage{url}

\usepackage{graphicx}
\usepackage{amsmath}
\usepackage{amssymb}
\usepackage{booktabs}
\usepackage{enumitem}
\usepackage{multirow}
\usepackage{wrapfig}
\usepackage{makecell}
\usepackage[accsupp]{axessibility}
\usepackage{footnote}
\usepackage{comment}
\usepackage{algorithm}
\usepackage{algpseudocode}
\usepackage{footmisc}
\usepackage{marvosym}
\usepackage{fontawesome5}

\usepackage[noend]{algcompatible}
\usepackage{arydshln} 
\usepackage[
  separate-uncertainty = true,
  multi-part-units = repeat
]{siunitx}

\usepackage[capitalize]{cleveref}
\crefname{section}{Sec.}{Secs.}
\Crefname{section}{Section}{Sections}
\Crefname{table}{Table}{Tables}
\crefname{table}{Tab.}{Tabs.}

\def\paperID{} 
\def\confName{CVPR}
\def\confYear{2026}

\title{ConLA: Contrastive Latent Action Learning from Human Videos for Robotic Manipulation}

\author{
Weisheng Dai$^{1\dagger}$ \quad
Kai Lan$^{2,3\dagger}$ \quad
Jianyi Zhou$^{1}$ \quad
Bo Zhao$^{4}$\\[3pt]
Xiu Su$^{5}$ \quad
Junwen Tong$^{2,3}$ \quad
Weili Guan$^{1}$ \quad
Shuo Yang$^{1}$\textsuperscript{\Letter} \\[3pt]
$^{1}$ Harbin Institute of Technology, Shenzhen \\
$^{2}$ State Key Laboratory of Mobile Network and Mobile Multimedia Technology, Shenzhen\\\quad
$^{3}$ ZTE Corporation\quad
$^{4}$ Shanghai Jiao Tong University \quad
$^{5}$ Central South University\\[3pt]
{\tt\small shuoyang@hit.edu.cn}}

\begin{document}

\twocolumn[{%
\renewcommand\twocolumn[1][]{#1}%
   \maketitle
    \begin{center}
       \vspace{-1pt} %
        \centering
        \includegraphics[width=0.80\linewidth]{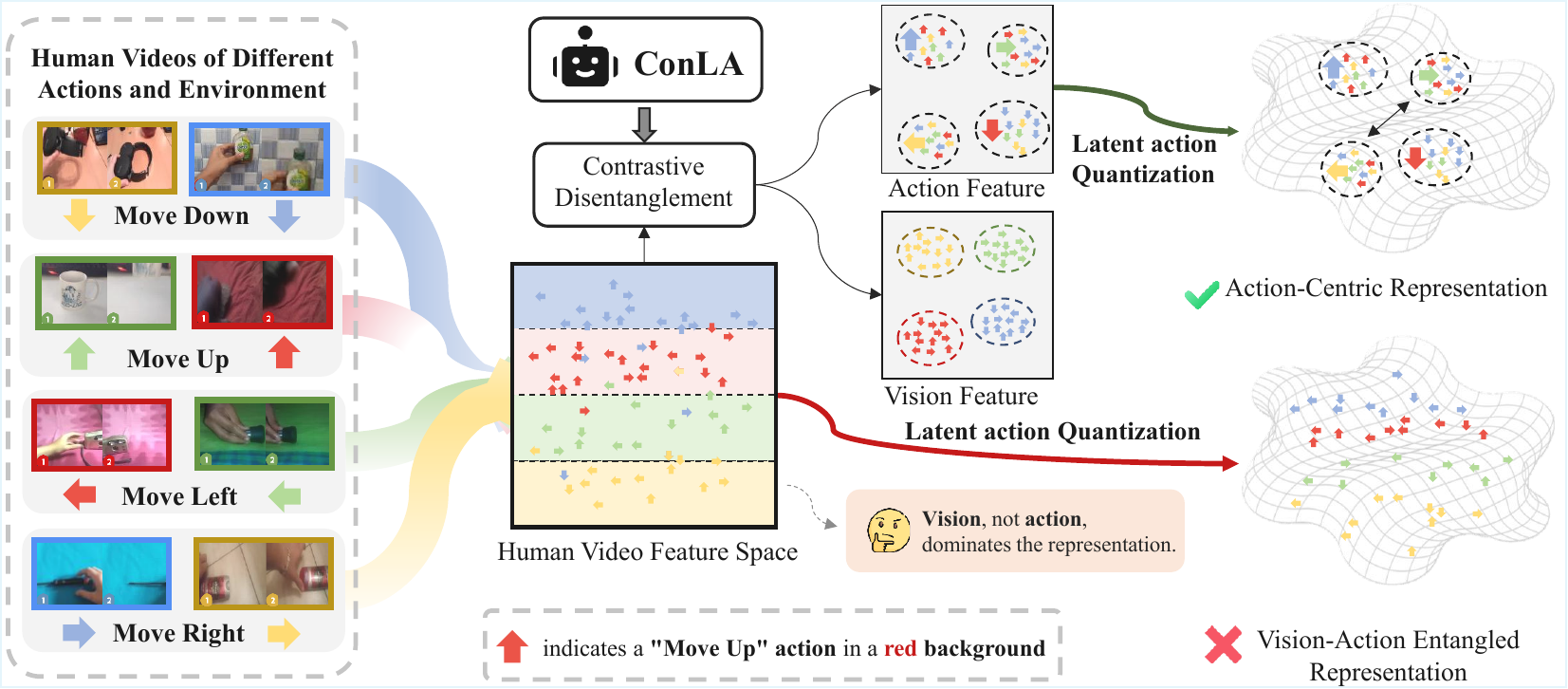}
        \captionof{figure}{The overview of ConLA, which leverages contrastive learning to disentangle latent actions from visual noise in videos, guiding the construction of compact latent action representations. This enables the model to learn motion priors from complex human videos, improving downstream robot manipulation tasks.}
        \label{fig:start}
    \end{center}
}]

\begingroup
\makeatletter
\renewcommand\thefootnote{}%
\setlength{\skip\footins}{2pt}
\footnotetext{%
\footnotesize
\Letter\ Corresponding author \quad
$^{\dagger}$ Equal contribution
}
\endgroup
\setcounter{footnote}{0}

\input{sec/0_abstract}    
\input{sec/1_intro}
\input{sec/2_Related_Works}

\input{sec/3_Method}
\input{sec/4_exp}

\input{sec/5_Summary}

\clearpage
{\small
\bibliographystyle{ieeenat_fullname}
\bibliography{reference}}

\clearpage
\appendix
\input{sec/6_Supplementary}
\end{document}

%% file: sec/0_abstract.tex
\begin{abstract}
Vision-Language-Action (VLA) models achieve preliminary generalization through pretraining on large scale robot teleoperation datasets. However, acquiring datasets that comprehensively cover diverse tasks and environments is extremely costly and difficult to scale.
In contrast, human demonstration videos offer a rich and scalable source of diverse scenes and manipulation behaviors, yet their lack of explicit action supervision hinders direct utilization.
Prior work leverages VQ-VAE based frameworks to learn latent actions from human videos in an unsupervised manner. Nevertheless, since the training objective primarily focuses on reconstructing visual appearances rather than capturing inter-frame dynamics, the learned representations tend to rely on spurious visual cues, leading to shortcut learning and entangled latent representations that hinder transferability. To address this, we propose ConLA, an unsupervised pretraining framework for learning robotic policies from human videos. ConLA introduces a contrastive disentanglement mechanism that leverages action category priors and temporal cues to isolate motion dynamics from visual content, effectively mitigating shortcut learning. Extensive experiments show that ConLA achieves strong performance across diverse benchmarks. Notably, by pretraining solely on human videos, our method for the first time surpasses the performance obtained with real robot trajectory pretraining, highlighting its ability to extract pure and semantically consistent latent action representations for scalable robot learning. Our code and
data are available at \url{https://github.com/WeishengDAI/ConLA}


\end{abstract}

%% file: sec/1_intro.tex
\section{Introduction}
\label{sec:intro}

Recent advances in large language models (LLMs) have revealed predictable scaling laws: as model size, dataset scale, and computation increase, performance improves and generalization emerges naturally~\cite{kaplan2020scaling,srivastava2023beyond,wei2022emergent,achiam2023gpt}. Inspired by this, Vision-Language-Action (VLA)~\cite{black2410pi0,kim2025openvla,driess2023palm,liu2024rdt,wen2025dexvla,wen2025tinyvla,pertsch2025fast,bjorck2025gr00t} models have shown promising progress by pre-training on large-scale robotic teleoperation data, achieving preliminary generalization. However, acquiring robot teleoperation datasets that both cover all possible environments and encompass diverse tasks is practically infeasible, and for certain specific environments or tasks, data collection can be extremely challenging or even impossible. By contrast, the vast abundance of human demonstration videos offers a naturally rich and scalable data source for VLA models, with significant potential to enhance generalization. Nevertheless, these videos lack explicit robotic action trajectories, making direct VLA training challenging.



\begin{figure}[t!]
\centering
   \includegraphics[width=0.85\linewidth]{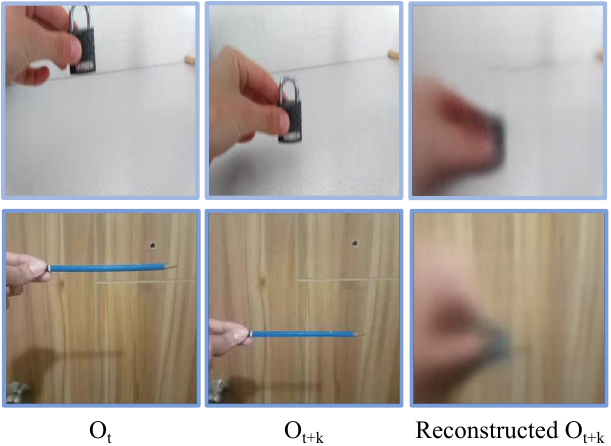}
\caption{Illustration of shortcut learning: using the latent action extracted from the first-row frame pair to reconstruct the second-row $O_{t+k}$ fails, as the reconstruction drives the model to capture appearance rather than motion.}
\label{fig:meth_motivation}
\end{figure}

To address this challenge, a line of recent work has emerged~\cite{yelatent,bjorck2025gr00t,bu2025univla,chen2024moto}. LAPA~\cite{yelatent} first introduced the idea of leveraging unlabeled videos for latent action learning to pretrain VLA models, extracting latent action from videos using a VQ-VAE~\cite{van2017neural} paradigm to transfer human video motion prior into VLA models. While promising, these approaches suffer from a fundamental limitation: VQ-VAE~\cite{van2017neural} based latent action extraction methods are prone to shortcut learning, as the vision reconstruction-based optimization objective provides no direct incentive for learning meaningful latent action. As a result, as shown in Fig.~\ref{fig:meth_motivation}, the model often fails to capture meaningful motion information and instead memorizes future visual content to minimize reconstruction error, thereby producing an entangled action space mixed with irrelevant visual features. 
This problem is particularly pronounced in human videos, as their inherent complex visual variations make latent action extraction more difficult and limit their transferability to robot learning. This raises a crucial question: can we mitigate the impact of shortcut learning and extract more pure latent action from human videos to unlock the full potential of human video pretraining for VLA models?
In unsupervised settings, disentangling motion from mixed visual and motion cues is challenging~\cite{locatello2019challenging}, necessitating explicit priors to guide the extraction of meaningful latent action. We observe that human manipulation videos consist of a large number of recurring action primitives (e.g., picking, placing, moving), which provide natural semantic cues for latent action learning. Building on these natural semantic cues, leveraging action category information as a supervisory signal encourages \textit{\textbf{latent actions of the same category to cluster compactly across different environments and embodiments}}, enhancing their semantic consistency and preventing the model from memorizing visual content to minimize reconstruction loss. Moreover, videos naturally contain rich temporal information: motion features are highly sensitive to temporal order, whereas visual appearance remains relatively stable. By exploiting temporal prior, the model can separate motion dynamics from static visual cues, achieving a more effective disentanglement of motion and appearance in the latent action space. 

Building upon these insights, we propose ConLA, an unsupervised pretraining framework for robotic policy learning from human videos. We aim to extract a compact and semantically consistent latent action representation from human demonstrations to facilitate the transfer of motion knowledge to robot learning. ConLA consists of three key stages: 1) \textbf{Contrastive Latent Action Learning}, where we leverage action category priors and temporal priors in videos to disentangle latent actions from mixed visual noise through contrastive learning. 
Specifically, we first extract latent action representations from paired frames by modeling inverse dynamics with a VQ-VAE~\cite{van2017neural}. Before discretization, these representations are processed by our contrastive disentanglement module, which employs contrastive learning to guide the latent action embeddings, effectively isolating pure and compact latent actions. 2) \textbf{Latent Action Pretraining}, where we leverage the discretized and semantically consistent latent action tokens obtained from the first stage to train an auto-regressive vision-language model. The model predicts latent actions from video observations and task instructions, enabling the transfer of human motions from video demonstrations to robot policies. 3) \textbf{Action Finetuning}, we finetune the model using a small amount of real robot data, mapping latent actions to executable motor actions to obtain the final policy.
On multiple benchmarks, ConLA consistently achieves state-of-the-art performance. Compared with our baseline LAPA~\cite{yelatent}, ConLA shows significant improvements. By effectively extracting semantically meaningful latent actions, ConLA pretrained solely on human videos improves over LAPA~\cite{yelatent} by 12.5\% on the SimplerEnv~\cite{li2025evaluating} benchmark, and notably, even exceeds the performance of models pretrained directly on real robot trajectories by 1.1\%. These exciting results demonstrate the feasibility and potential of large-scale human videos for VLA training. ConLA successfully extracts high-quality latent action and validates the effectiveness of knowledge transfer from human demonstrations. Our contributions are:
\begin{itemize}
    \item 
    We identify that existing VQ-VAE~\cite{van2017neural} based latent action learning methods suffer from shortcut learning, where models rely excessively on visual appearance cues rather than modeling true motion dynamics. To address this, we introduce contrastive learning to disentangle visual and action representations, enabling latent actions that more faithfully capture real motion semantics.
    \item 
    We propose a contrastive disentanglement architecture that leverages action category and temporal priors to ensure that latent actions with the same semantics cluster compactly across environments and embodiments, improving latent action learning from human videos. 
    \item ConLA achieves state-of-the-art performance on both simulation benchmarks and real-robot tests, achieving an \textbf{12.5\%} increase in success rate over LAPA~\cite{yelatent} on the SimplerEnv benchmark~\cite{li2025evaluating}, and a \textbf{15.9\%} improvement in real-world tests. Moreover, the policy pretrained on human videos exceeds those trained with robot trajectory data by 1.1\%, demonstrating the feasibility of scaling VLA using large-scale human video datasets.
\end{itemize}

\begin{figure*}[t!]
\setlength{\abovecaptionskip}{ 2 pt}
\setlength{\belowcaptionskip}{ -6 pt}
\centering
   \includegraphics[width=1.\linewidth]{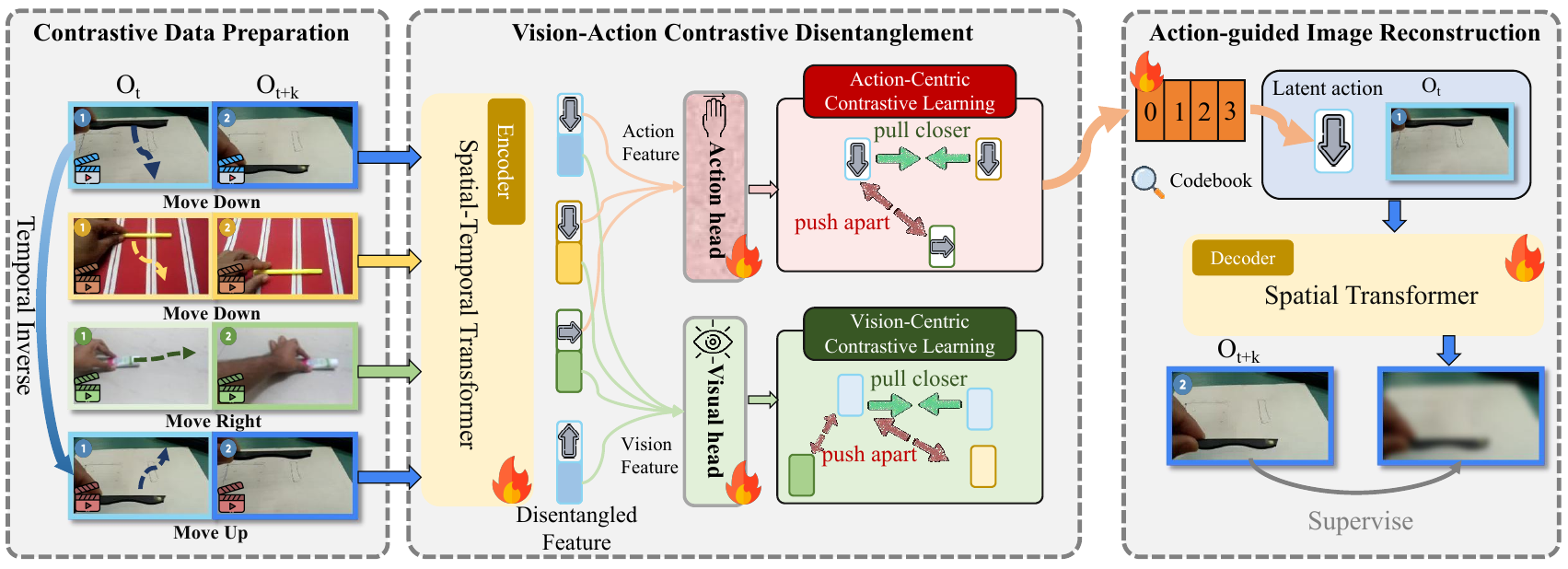}
\caption{Contrastive Latent Action Learning. We propose a contrastive disentanglement framework to separate action from visual interference in video clips spanning the current and future frames. Specifically, samples with action class labels and their inversely augmented counterparts are encoded into latent action embeddings, which are evenly divided and fed into the Action head for Action-Centric Contrastive Learning and the Visual head for Vision-Centric Contrastive Learning to achieve disentangled representations. The optimized representation from the Action head is further quantized, and the resulting quantized latent actions, together with the current frame $O_t$, are employed to reconstruct the future frame $O_{t+k}$.}
\label{fig:meth_framework}
\end{figure*}

%% file: sec/2_Related_Works.tex
\section{Related Works}
\label{sec:Related Works}

\noindent\textbf{Vision-Language-Action Models.} Building upon the success of large language models (LLMs) and vision-language models (VLMs)~\cite{zhu2025internvl3,xu2025qwen2,lu2024deepseek,achiam2023gpt,team2024gemini,touvron2023llama,anil2023palm}, researchers have recently introduced Vision-Language-Action Models (VLAs)~\cite{zitkovich2023rt,bjorck2025gr00t,team2024octo,kim2025openvla,black2410pi0,pertsch2025fast,brohan2022rt,liu2024rdt,li2024cogact,liu2025hybridvla}. These models map visual observations and language instructions into robotic actions, enabling the execution of manipulation tasks. OpenVLA~\cite{kim2025openvla} pretrains on large-scale teleoperation datasets and models actions as tokens within the language model’s vocabulary, achieving generalist manipulation capabilities. $\pi$0~\cite{black2410pi0} and $\pi$0.5~\cite{intelligence2504pi0} further leverage cross-embodiment, multi-source teleoperation data and adopt a flow-matching~\cite{lipman2022flow} based architecture, which enhances the ability to perform fine-grained tasks and demonstrates stronger generalization. Despite these advances, existing approaches heavily rely on large-scale teleoperation datasets with action annotations, which constrains their scalability and limits broader applicability.\\

\noindent\textbf{Learning from Human Videos.} In real-world robot manipulation, collecting large-scale teleoperation data~\cite{bu2025agibot,vuong2023open,khazatsky2024droid} is difficult to scale. Consequently, learning from video demonstrations has emerged as a promising paradigm.

Some studies~\cite{kareer2025egomimic,liu2025egozero,yang2025egovla,qiu2025humanoid,luo2025being,zhou2025you,qin2022dexmv,lepert2025phantom,hoque2025egodex,chen2025fmimic,bahety2024screwmimic} attempt to explicitly extract structured information from human videos to facilitate robot learning. These approaches typically rely on hand pose estimators or motion capture systems to retarget human actions to the robot action space. EgoMimic~\cite{kareer2025egomimic} and HAT~\cite{qiu2025humanoid} train task-specific policies from egocentric human videos, but they depend on paired human-robot data, limiting scalability and generalization. Methods such as EgoVLA~\cite{yang2025egovla} and Being-H0~\cite{luo2025being} pretrain policies using egocentric human videos and achieve encouraging results; however, they still cannot leverage large-scale free Internet videos, require carefully collected human demonstrations, and must handle human-to-robot hand retargeting. Although more accessible than teleoperation data, these methods remain constrained by the effort required for data collection, limiting their scalability.

Another line of work~\cite{yelatent,liang2025clam,yang2025learning,bu2025univla,chen2024moto,kim2025uniskill,chen2025villa,tharwat2025latent} focuses on learning latent actions from videos and using them for policy modeling. These approaches typically rely on unsupervised inverse dynamics models (IDMs) to extract action priors from unlabeled videos, which are used to train VLA policies. LAPA~\cite{yelatent} leverages VQ-VAE~\cite{van2017neural} to extract motion priors from consecutive frames, enabling knowledge transfer from human videos to robot manipulation. CLAM~\cite{liang2025clam} and COMO~\cite{yang2025learning} highlight the limitations of discrete latent actions in terms of expressivity and advocate modeling latent actions in continuous action spaces to improve representation capacity. These approaches do not require external models or sensors, enabling large-scale use of Internet videos. However, VQ-VAE~\cite{van2017neural} based latent action extraction is prone to shortcut learning. UniVLA~\cite{bu2025univla} partially addresses this by reconstructing DINOv2~\cite{oquab2024dinov2} features of future frames and construct task-centric latent actions, reducing irrelevant environmental noise. Nevertheless, it lacks explicit inductive biases, and its representations still fail to fully capture motion semantics in human videos. In contrast, our approach leverages intrinsic video priors, including action category and temporal cues, to guide the model in learning compact, disentangled representations that more effectively capture inter-frame dynamics while suppressing irrelevant visual distractions.

%% file: sec/3_Method.tex
\section{Methodology}
Our framework consists of three stages. In the first stage, we leverage action category information and the temporal cues in videos as inductive biases to extract latent actions from videos, thereby obtaining a set of discretized, semantically consistent latent actions. In the second stage, we pretrain an autoregressive VLM-based policy that predicts discrete latent action tokens given visual observations and task instructions. Finally, in the third stage, we fine-tune the policy on a small amount of real robot trajectories, establishing a mapping from latent actions to executable control signals.

\subsection{Contrastive Latent Action Learning}\label{Latent Action Learning}
In the first stage, we train a base model to generate pseudo labels (latent action tokens) for videos. Specifically, contrastive learning is employed to guide the disentanglement of latent action representations from visual noise, yielding more discriminative pseudo labels that serve as a reliable basis for policy pretraining in the second stage.

\noindent\textbf{Latent action quantization.} 
We construct a video pair $[O_t, O_{t+k}]$ from a current frame $O_t$ and a future frame $O_{t+k}$ with a frame interval of $k$, along with its corresponding action class label $y$. To incorporate a temporal prior, we apply a reverse-order augmentation to create the inverse pair $[O_{t+k}, O_t]$. Our latent action model consists of an Inverse Dynamics Model as encoder $I$ and a Forward Dynamics Model as decoder $F$. Following C-ViViT tokenizer~\cite{villegas2022phenaki}, our encoder is implemented as a spatial-temporal Transformer~\cite{xu2020spatial}, which takes the current frame $O_t$ and the future frame $O_{t+k}$ as input and extracts the motion information between the two frames, producing a latent action embedding $\boldsymbol Z \in\mathbb{R}^{d}$, with predefined dimension $d$. To obtain a semantically consistent latent action representation, $\boldsymbol Z$ is further processed by a Contrastive Disentanglement Module, resulting in a more discriminative and structured embedding $\boldsymbol Z_a$. We then apply latent quantization to $\boldsymbol Z_a$ to obtain $\boldsymbol Z_{aq}$, which is optimized using the VQ-VAE~\cite{van2017neural} objective with a codebook of size $|C|$. The decoder, implemented as a spatial Transformer, takes the current frame $O_t$ and the quantized latent action tokens $\boldsymbol Z_{aq}$ as input to generate the predicted future frame $\hat{O}_{t+k}$. Our objective minimizes the reconstruction error: $\| \hat{O}_{t+k} - O_{t+k} \|^2$.

\noindent\textbf{Contrastive Disentanglement Module.}
As illustrated in Figure \ref{fig:meth_motivation}, the prior paradigm based on the VQ-VAE~\cite{van2017neural} suffers from a pronounced shortcut learning problem: the latent actions learned by the model often encode discretized copies of the future frame’s visual content rather than the true inter-frame dynamics.
To mitigate the interference of visual information in latent action extraction, we introduce a contrastive disentanglement framework that incorporates Action-Centric Contrastive Learning and Vision-Centric Contrastive Learning. These two components jointly disentangle action from visual content, enabling the model to produce high-quality and semantically consistent latent action representations for downstream policy learning.
\noindent\textbf{1) Action-Centric Contrastive Learning:}
As illustrated in Figure \ref{fig:meth_framework}, after obtaining the latent action representation via the encoder, we evenly split $\boldsymbol Z$ into two parts: $\boldsymbol Z_{a^{\prime}}$ (action-related) and $\boldsymbol Z_{v^{\prime}}$ (visual-related) as follows:
\setlength{\abovedisplayskip}{3pt}
\setlength{\belowdisplayskip}{3pt}
\begin{align}
\boldsymbol Z &= I([O_t, O_{t+k}]), \quad \boldsymbol Z \in \mathbb{R}^{d} \\
\boldsymbol Z &= [\, \boldsymbol Z_{a^{\prime}} \,;\, \boldsymbol Z_{v^{\prime}} \,], \quad 
\boldsymbol Z_{a^{\prime}}, \boldsymbol Z_{v^{\prime}} \in \mathbb{R}^{d/2}
\end{align}
For $\boldsymbol Z_{a^{\prime}}$, we apply a two-layer MLP as an action head to project the representation into the action space, resulting in $\boldsymbol Z_{a}$. To learn compact latent action representations, we employ action-centric contrastive learning by optimizing an action loss, denoted as ${L}_{\text{action}}$, which is implemented as a supervised contrastive objective~\cite{khosla2020supervised}. Compared with LAPA~\cite{yelatent}, which relies solely on an unsupervised VQ-VAE~\cite{van2017neural}, incorporating weak supervision in the form of action class labels substantially improves the discriminability of the latent representations. Without supervision, latent actions are highly susceptible to visual distractions (e.g., background variations), which may cause similar actions to be encoded as entirely different latent representation, leading to a entangled representation space. The action loss mitigates this by pulling representations of the same action class closer while pushing apart those of different classes, leading to a compact and semantically coherent clustering in the latent space. This mechanism effectively alleviates shortcut learning and yields more discriminative latent action representations. This process can be describes as:
\begin{equation}
\boldsymbol Z_{a} =\text{MLP}_{\text{action}}(\boldsymbol Z_{a^{\prime}}), \quad 
\boldsymbol Z_{a}\in \mathbb{R}^{d}
\end{equation}

\begin{equation}\label{eq:action loss}
\mathcal{L}_{\text{action}} =
\sum_{i \in I} \frac{-1}{|P(i)|} 
\sum_{p \in P(i)} 
\log 
\frac{
    \exp\left(\boldsymbol Z_{a,i} \cdot\boldsymbol Z_{a,p} / \tau\right)
}{
    \sum\limits_{a \in A(i)} \exp\left(\boldsymbol Z_{a,i} \cdot \boldsymbol Z_{a,a}/ \tau\right)
},
\end{equation}
where, $i \in I \equiv \{1, \dots, N\}$ denotes the index of a sample, referred to as the anchor. $\boldsymbol Z_{a,i}$ represents the action embedding of the $i$-th sample, $\tau$ is a scalar temperature parameter, and $A(i) \equiv I \setminus \{i\}$ denotes the set of all indices in the batch excluding $i$. The set $\{\, p \in A(i) : \tilde{y}_p = \tilde{y}_i \,\}$  represents all positive samples for the anchor $i$ (i.e., samples sharing the same action label as $i$), and $\lvert P(i) \rvert$ denotes its cardinality.

\noindent\textbf{2) Vision-Centric Contrastive Learning:}
In inverse dynamics modeling, the difference between the current and future frames contains not only motion information but also unavoidable environmental noise, such as camera jitter, viewpoint changes, or illumination fluctuations. Without inductive biases, it is challenging to disentangle these components using unsupervised learning. We leverage the temporal sensitivity prior: when the frame order is reversed, motion information changes significantly, whereas content information and visual distractions remain relatively stable. Based on this prior, we introduce a Vision-Centric contrastive learning objective to maintain content consistency while reducing the influence of motion variations. Specifically, we take the reversed frame pair $[O_{t+k}, O_t]$ and pass it through the encoder to obtain the latent action representation for the inverse sequence, denoted as $\boldsymbol Z^{I}$. We then split $\boldsymbol Z^{I}$ evenly into two parts, yielding $\boldsymbol Z_{a^{\prime}}^I$ and $\boldsymbol Z_{v^{\prime}}^I$ as follows:
\setlength{\abovedisplayskip}{3pt}
\setlength{\belowdisplayskip}{3pt}
\begin{align}
\boldsymbol Z^{I} &= I([O_{t+k}, O_{t}]), \quad \boldsymbol Z^{I} \in \mathbb{R}^{d} \\
\boldsymbol Z^{I} &= [\, \boldsymbol Z_{a^{\prime}}^{I} \,;\, \boldsymbol Z_{v^{\prime}}^{I} \,], \quad \boldsymbol Z_{a^{\prime}}^{I}, \boldsymbol Z_{v^{\prime}}^{I} \in \mathbb{R}^{d/2}
\end{align}
We project $\boldsymbol Z_{v^{\prime}}$ and $\boldsymbol Z_{v^{\prime}}^I$ into the visual space through the visual head, yielding $\boldsymbol Z_v$ and $\boldsymbol Z_v^I$ as follows:
\begin{equation}
\setlength{\abovedisplayskip}{3pt}
\setlength{\belowdisplayskip}{3pt}
\begin{aligned}
\boldsymbol Z_{v} &= \text{MLP}_{\text{visual}}(\boldsymbol Z_{v^{\prime}}), \quad \boldsymbol Z_{v} \in \mathbb{R}^{d} \\
\boldsymbol Z_{v}^{I} &= \text{MLP}_{\text{visual}}(\boldsymbol Z_{v^{\prime}}^{I}), \quad \boldsymbol Z_{v}^{I} \in \mathbb{R}^{d}
\end{aligned}
\end{equation}
We treat the inverse visual representation $\boldsymbol Z_{v}^{I}$ as a positive sample to construct a Vision-Centric Contrastive Learning objective, where the optimization is guided by a visual loss
$\mathcal{L}_{\text{visual}}$ implemented as an InfoNCE~\cite{chen2020simple} loss. The vision-centric contrastive objective encourages the model to capture content-consistent and motion-invariant features. By contrasting visual representations under motion perturbations, the visual loss drives the model to isolate appearance information from dynamic changes, thereby promoting the disentanglement of visual and motion representations.
The formulas are expressed as follows:
\begin{equation}\label{eq:visual loss}
\mathcal{L}_{\text{visual}} = - \sum_{i \in I}
\log 
\frac{ \exp(\boldsymbol{\tilde{Z}}_{v,i} \cdot \boldsymbol{\tilde{Z}}_{v,j(i)} / \tau) }
     { \sum\limits_{a\in A(i)} \exp( \boldsymbol{\tilde{Z}}_{v,i} \cdot \boldsymbol{\tilde{Z}}_{v,a}/ \tau) }.
\end{equation}
Here, $i \in I \equiv \{1, \dots, 2N\}$ denotes the index of a sample, and let $j(i)$ be the index of the positive sample corresponding to anchor sample $i$. $\boldsymbol{\tilde{Z}}_v= [\boldsymbol{Z}_v; \boldsymbol{Z}_v^I] \in \mathbb{R}^{2N \times d}$ denotes the concatenated visual embeddings of a batch containing $2N$ samples, where $\boldsymbol{Z}_v \in \mathbb{R}^{N \times d}$ and $\boldsymbol{Z}_v^I \in \mathbb{R}^{N \times d}$ serve as positive pairs of each other. $\boldsymbol{\tilde{Z}}_{v,i}$ denotes the visual embedding of the $i$-th sample in the batch.

\subsection{Latent Action Pretraining}\label{Latent Action Pretraining}
We leverage the latent action quantization encoder trained in the first stage as an inverse dynamics model to extract latent actions from videos, which serve as pseudo-labels. Specifically, for each pair of current frame $O_t$ and future frame $O_{t+k}$, we generate the corresponding latent action by retrieving the nearest quantized representation from the action-centric codebook, thereby constructing a dataset consisting of observation--instruction--pseudo action label triplets. We then perform latent action pretraining on this dataset by employing a pretrained vision-language model (VLM) to predict $Z_{aq}$ conditioned on the task instruction and the current frame $O_t$. Following LAPA~\cite{yelatent}, we attach an additional latent action head after the language model head of the VLM, implemented as a single-layer MLP with vocabulary size $|C|$. During training, the vision encoder is frozen while the language model is unfrozen for optimization. To maintain consistency, our generalist policy is based on the 7B Large World Model~\cite{liu2024world}.

\subsection{Action Finetuning}\label{Action Finetuning}
After the second stage of latent action pretraining, the motion priors from videos have been successfully transferred to the policy. However, the resulting latent actions cannot be directly executed on downstream robotic tasks, as they do not correspond to actual end-effector movements. To map latent actions to real robot actions, we finetune the pretrained policy using a small set of trajectories containing ground-truth robot actions. During action prediction, we discretize the continuous action space of each robot dimension. In the finetuning stage, the original latent action head is discarded and replaced with a new action head to generate ground-truth actions. Consistent with latent action pretraining, the vision encoder is frozen while all parameters of the underlying language model are unfrozen for optimization.

%% file: sec/4_exp.tex
\section{Experiments}
\label{sec:Experiments}
To demonstrate the effectiveness of our proposed generalist policy, our framework is evaluated in both the SimplerEnv~\cite{li2025evaluating} simulation environment and real-world scenarios. In addition, we conduct latent action analysis and perform ablation studies to investigate critical design choices.

\subsection{Benchmarks}\label{Benchmarks}
\textbf{SimplerEnv}~\cite{li2025evaluating} is designed to faithfully reflect the performance of real-world policies by mirroring physical dynamics and visual appearances. We focus on four tasks in the “WidowX + Bridge” setup: (1) putting a spoon on a table cloth, (2) placing a carrot on a plate, (3) stacking a green cube on a yellow cube, and (4) putting an eggplant into a basket. Since SimplerEnv~\cite{li2025evaluating} lacks fine-tuning trajectories, we follow the experimental setup of LAPA~\cite{yelatent} and collect 100 multi-task trajectories based on successful rollouts from a VLA model trained on the BridgeV2 dataset~\cite{walke2023bridgedata}. The pose and position of the objects to be grasped are randomly initialized using different seeds. Our evaluation follows LAPA~\cite{yelatent}, with each task assessed over 24 independent trials to ensure robust performance metrics.\\

\noindent\textbf{Real-World Tabletop Manipulation} experiments are conducted using a 7-DoF Franka Research 3 robot arm in three environments, equipped with a third-view Realsense D435i RGB-D camera, from which we use only RGB images. We leverage two pretrained data sources, BridgeV2~\cite{walke2023bridgedata} and Something-SomethingV2~\cite{goyal2017something}, following the real-world experimental setup of LAPA~\cite{yelatent}. The model is fine-tuned on three multi-instruction tasks: (1) Knock \textless object\textgreater{} Over, (2) Cover \textless object\textgreater{} with Towel, (3) Pick \textless object\textgreater{} into Box. For each task, we collect 150 trajectories. For evaluation, we adopt a task-specific partial success criterion, following OpenVLA~\cite{kim2025openvla}. More details are in the Appendix.

\subsection{Pretraining Datasets}\label{Pretraining Datasets}
We pretrain VLM policy on both a robot video dataset and a human video dataset. \textbf{1) BridgeV2}~\cite{walke2023bridgedata} is a large-scale robotic manipulation dataset containing 60,096 trajectories across 24 environments. The dataset encompasses a variety of skills, including picking, placing, pushing, sweeping, stacking, and folding. All trajectories are paired with natural language instructions. For data preprocessing, we categorized the language instructions into 80 action classes, forming the action class labels used in the first-stage latent action learning. Details of the data preprocessing are provided in Appendix. \textbf{2) Something-SomethingV2}~\cite{goyal2017something} is a collection of 220,847 labeled video clips of humans performing predefined, basic actions with everyday objects. Although the dataset does not contain ground-truth action labels, it provides predefined action class labels for each video clip, covering a total of 174 action categories. 

\subsection{Baselines}\label{Baselines}
Our selected baseline models include the following models:
\textbf{1) UNIPI}~\cite{du2023learning} adopts a video diffusion model for language-conditioned rollout generation during pretraining and employs an inverse dynamics model for finetuning on real actions.
\textbf{2) VPT}~\cite{baker2022video} trains an inverse dynamics model on labeled data to extract pseudo actions from videos, which are then used to pretrain a VLM.
\textbf{3) LAPA}~\cite{yelatent} learns latent actions from videos using a naive VQ-VAE~\cite{van2017neural}, and leverages the extracted latent actions to pretrain VLM.
\textbf{4) SCRATCH} denotes training the same backbone VLM from scratch on the fine-tuning dataset only, serving as a lower-bound baseline to assess pretraining gains.
\textbf{5) ACTIONVLA} denotes pretrains the same backbone VLM using ground-truth robot action data, which can be regarded as an upper bound since it relies on access to real action labels.

\subsection{Evaluation on SimplerEnv}\label{SimplerEnv Results}
In this section, we pretrain policies using both robot videos (BridgeV2~\cite{walke2023bridgedata}) and human videos (Something-SomethingV2~\cite{goyal2017something}). Robot videos are collected in controlled environments with minimal noise but are limited and expensive to obtain. In contrast, human videos are abundant and easy to access, yet contain high environmental noise that challenges latent action learning. This experiment evaluates the generality of our approach across both video types, and examines whether improved latent action representations can mitigate challenges in human videos, enhance their utility, and facilitate the transfer of motion priors to robot manipulation tasks.

\newcommand{\gain}[1]{\textcolor[rgb]{0.0,0.6,0.0}{\textbf{(#1)}}}
\begin{table*}[t]
\centering
\caption{Quantitative results of our method and baselines on the SimplerEnv. Average success rate (\%) are shown. Notably, ConLA pretrained only on human videos surpasses model pretrained on real robot trajectories (ACTIONVLA) by 1.1\%.}
\small
\renewcommand{\arraystretch}{0.9} 
\begin{tabular}{l l l c c c c l}
\toprule
\makecell[l]{Pretraining Data} & \makecell[l]{Data Type} & Policy & 
\makecell[c]{stack green \\ to yellow block} & 
\makecell[c]{put carrot \\ on plate} & 
\makecell[c]{put spoon \\ on towel} & 
\makecell[c]{put eggplant \\ in basket} & 
Average \\ 
\midrule
\makecell[c]{-} & \makecell[c]{-} & SCRATCH & 29.2 & 29.2 & 50.0 & 29.2 & 34.4\\
BridgeV2~\cite{walke2023bridgedata} & Robot Trajectories & ACTIONVLA & 75.0 & 58.0 & 70.8 & 50.0 & 63.5 \\ 
\midrule
\multirow{4}{*}{BridgeV2~\cite{walke2023bridgedata}} 
& \multirow{4}{*}{Robot Videos} & UNIPI~\cite{du2023learning} & 2.7 & 2.7 & 0.0 & 0.0 & 1.3 \\
&  & VPT~\cite{baker2022video} & 45.8 & 37.5 & 70.8 & 50.0 & 51.0 \\
&  & LAPA~\cite{yelatent} & 54.2 & 45.8 & 70.8 & 58.3 & 57.3 \\ 
&  & \textbf{ConLA (ours)} & \textbf{62.5} & \textbf{45.8} & \textbf{75.0} & \textbf{58.3} & \textbf{60.4} \gain{+3.1} \\ 
\midrule
\multirow{4}{*}{SomethingV2~\cite{goyal2017something}} 
& \multirow{4}{*}{Human Videos} & UNIPI~\cite{du2023learning} & 0.0 & 1.3 & 1.3 & 0.0 & 0.7 \\
&  & VPT~\cite{baker2022video} & 50.0 & 29.1 & 37.5 & 66.6 & 45.8 \\
&  & LAPA~\cite{yelatent} & 50.0 & 50.0 & 50.0 & 58.3 & 52.1 \\ 
&  & \textbf{ConLA (ours)} & \textbf{62.5} & \textbf{50.0} & \textbf{79.2} & \textbf{66.6} & \textbf{64.6} \gain{+12.5} \\ 
\bottomrule
\end{tabular}
\label{tab:task_results}
\end{table*}

\textbf{Results \& Analysis.}
As shown in Table~\ref{tab:task_results}, our method outperforms all baselines by large margins on SimplerEnv. Notably, ConLA pretrained solely on human videos even surpasses the model pretrained on real robot trajectories (ACTIONVLA). These results highlight that prior paradigms fail to address the challenges of human video data: despite being larger in scale than robot datasets and containing richer, more diverse motion information, human videos pretraining underperforms. By contrast, our method leverages human videos more efficiently, yielding promising results and paving the way for unlocking their full potential in future research. More results are in Appendix. 

\subsection{Real-World Results}\label{Real-World Results}
\begin{figure}[t]
\centering
\setlength{\abovecaptionskip}{2pt}
\setlength{\belowcaptionskip}{-10pt}
\includegraphics[width=\linewidth]{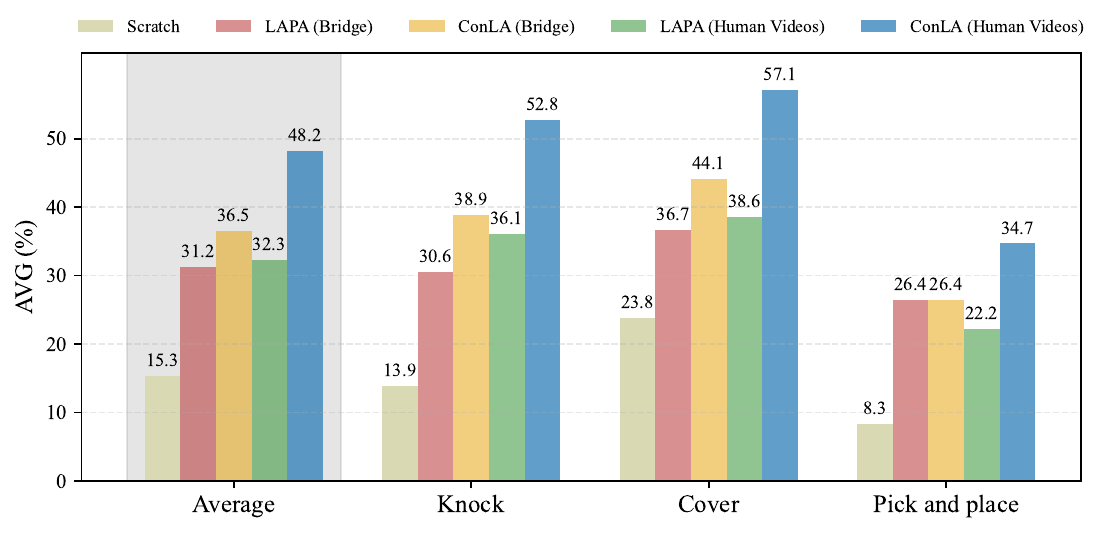}
\caption{Real-world Manipulation Robot Results. ConLA outperforms prior state-of-the-art by 15.9\% in success rate.}
\label{fig:result_bar}
\end{figure}

In this section, we pretrain policies using BridgeV2~\cite{walke2023bridgedata} and Something-SomethingV2~\cite{goyal2017something}, followed by finetuning with a small amount of real-world robot trajectories. Figure~\ref{fig:result_bar} reports real-robot performance across three tasks and three generalization settings (unseen object combination, unseen object, unseen instruction). Both LAPA~\cite{yelatent} and ConLA outperform SCRATCH, validating the value of video pretraining. However, LAPA~\cite{yelatent} shows almost no advantage when pretrained on human videos compared to robot videos, suggesting that despite the diversity and scale of human videos, domain complexity and distribution shift hinder effective use. In contrast, ConLA not only further improves BridgeV2~\cite{walke2023bridgedata} pretraining, but more importantly, achieves a significant performance boost when pretrained on human videos, surpassing LAPA~\cite{yelatent} by 15.9\%. We believe this improvement stems from ConLA’s ability to extract semantically consistent latent actions, enabling faithful motion prior acquisition from human videos. Additionally, as shown in Table~\ref{tab:realworld_results}, both LAPA~\cite{yelatent} and ConLA exhibit strong generalization particularly under the unseen object setting when pretrained on human videos due to the broader object diversity in large-scale human video datasets. Overall, these results highlight the scalability potential of human video pretraining and demonstrate that ConLA significantly enhances the transfer of human motion priors for downstream robot control. More results are in Appendix. 

\begin{table}[t]
\centering
\caption{Real-world generalization results across three evaluation settings. We report success rates (\%) on : (1) seen objects but unseen combinations, (2) unseen object, (3) new instructions requiring semantic reasoning.}
\resizebox{\linewidth}{!}{
\begin{tabular}{lcccc}
\toprule
Method & \makecell{Seen Obj.\\Unseen Combo} & Unseen Obj. & \makecell{Seen Obj.\\Unseen Instr.} & AVG \\
\midrule
SCRATCH        &18.4   &10.5   &17.1   &15.3\\
\midrule
LAPA (Bridge)  & 36.0  & 22.1   & 35.6  & 31.2 \\
ConLA (Bridge) & 46.2  & 25.4   & 37.8  & 36.5 \\
\midrule
LAPA (Human Videos) & 36.0 & 25.8 & 35.1 & 32.3 \\
ConLA (Human Videos)& \textbf{59.1} & \textbf{47.2} & \textbf{38.3} & \textbf{48.2} \\
\bottomrule
\end{tabular}
}
\label{tab:realworld_results}
\end{table}

\subsection{Analysis of Latent action}\label{Analysis of Latent action}
\begin{figure}[thb]
\centering
   \includegraphics[width=1.0\linewidth]{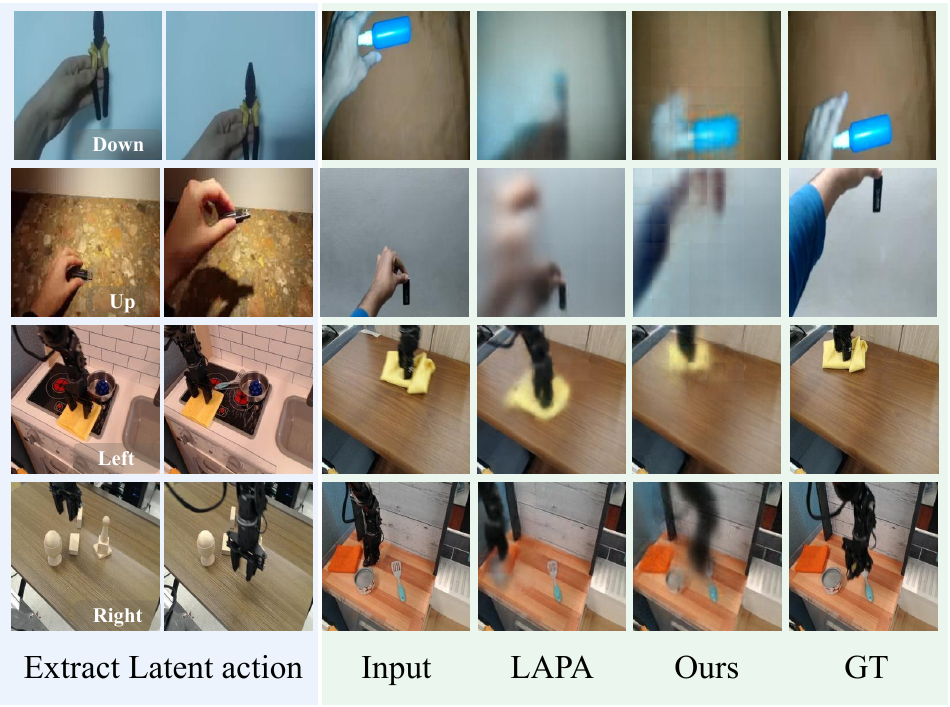}
\caption{Latent action analysis. Visualization of shortcut learning in latent action extraction. Reconstructed images conditioned on the extracted latent actions demonstrate that our method captures motion-relevant actions, alleviating shortcut learning.}
\label{fig:latent action reconstruction}
\end{figure}

\noindent\textbf{Shortcut Learning Analysis.} To assess the effectiveness of our method in mitigating shortcut learning during latent action extraction, we performed qualitative visualization analysis (Figure~\ref{fig:latent action reconstruction}). Specifically, we extracted four representative latent actions downward, upward, leftward, and rightward from a pair of video clips. These extracted latent actions were then applied to reconstruct frames from other images, aiming to assess whether the learned latent actions can control motion generation across different visual contexts. In this analysis, we used the current frame (input) as the condition and generated predicted future frames based on the extracted latent actions. We compared the reconstruction results of our method with those of a naïve LAPA~\cite{yelatent} baseline. As shown, in human videos, LAPA~\cite{yelatent} suffers from severe shortcut learning, where the extracted latent actions are dominated by visual content (as reflected in the right image of the first column) rather than motion semantics, whereas in robot videos, the extracted latent actions also exhibit semantic inconsistency. In contrast, our method captures motion-meaningful latent actions that truly represent underlying dynamics. These results demonstrate that our approach effectively mitigates shortcut learning and learns semantically consistent latent action representations.

\begin{figure}[h]
\centering
   \includegraphics[width=1.0\linewidth]{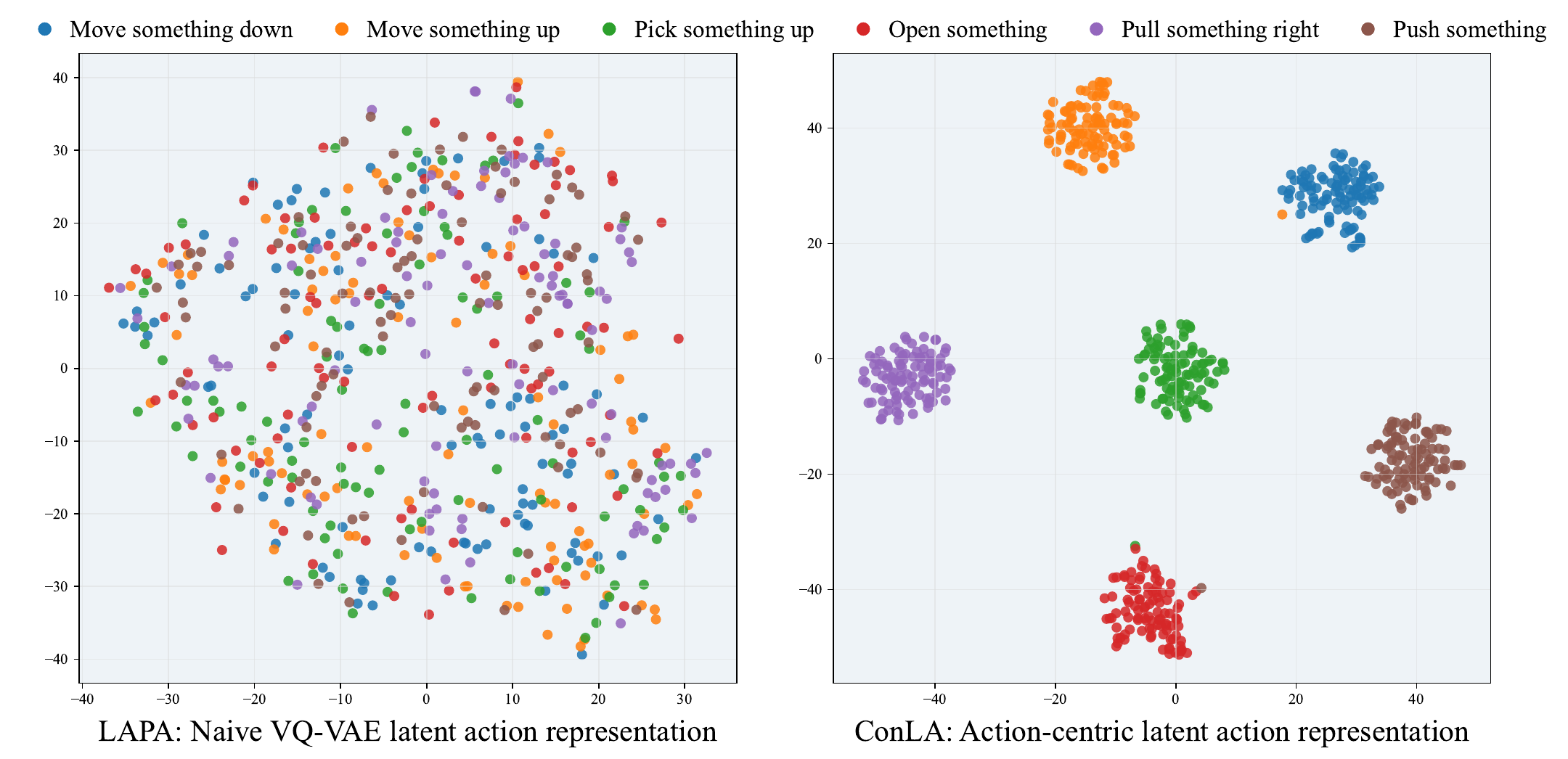}
\caption{t-SNE visualizations of the latent action embeddings show that our method yields semantically consistent and compact representations, with same-category actions forming tight clusters.}
\label{fig:latent action Space}
\end{figure}

\textbf{Latent Action Representation Analysis.} To analyze the structure of the latent action representation space, we randomly sample 100 video clips from each of action categories and extract their latent action embeddings, which are visualized using t-SNE. As shown in Figure~\ref{fig:latent action Space}, the latent action space obtained by the naïve VQ-VAE~\cite{van2017neural} is messy and entangled across different action categories (as shown in the left figure), whereas our method produces a more compact and semantically coherent latent action space. Similar motions sharing the same underlying dynamics are no longer separated due to differences in visual appearance. Such a representation enables a more faithful transfer of human motion priors to robotic training, thereby improving the efficiency of leveraging human video data for robot learning.

\subsection{Ablation Study}\label{Ablation Study}
\textbf{Contrastive Disentanglement Module.} To evaluate the contribution of each component in the disentanglement process, we conduct ablation studies on the first-stage latent action learning using the Something-SomethingV2~\cite{goyal2017something} dataset, and verify the performance on the SimplerEnv~\cite{li2025evaluating} benchmark with average task success rate as the evaluation metric. 
As shown in Table~\ref{tab:conla_ablation_avg}, using LAPA~\cite{yelatent} as the baseline, action-centric contrastive learning significantly improves latent action representations by leveraging weak supervision from action category labels. We further introduce vision-centric contrastive learning and examine the effect of inverse-order augmentation. Removing temporal inversion—feeding adjacent frames in their original order—causes action and visual embeddings to become more similar, leading to entangled representations and a performance drop. In contrast, inverse-order augmentation preserves a clear separation between action and visual features, yielding additional performance gains.

\begin{table}[thb]
\centering
\caption{Ablation study on the contrastive disentanglement module design on SimplerEnv. Average success rate (\%) are shown.}
\begin{tabular}{lc}
\toprule 
Method & Avg. \\
\midrule 
LAPA (base) & 52.1 \\
+ Action contrast & 58.4 \\
+ Action + Visual contrast (w/o inv. aug.) & 57.3 \\
Full ConLA & \textbf{64.6} \\
\bottomrule
\end{tabular}
\label{tab:conla_ablation_avg}
\end{table}

\noindent\textbf{Data scalability.} To evaluate the scaling ability of our method on human demonstration video datasets, we conducted experiments on the Something-SomethingV2~\cite{goyal2017something} dataset. Specifically, we pretrained our model using varying proportions of the dataset, ranging from 10\% to 100\%, to examine how performance scales with increasing data size. We also compared our results against LAPA~\cite{yelatent} baseline to assess relative improvements. The results show that performance scales positively with data size, and our method makes more efficient use of the data than the baseline.

\begin{table}[thb]
\centering
\caption{Ablation study of pretraining on different scales of human videos on SimplerEnv. Average success rate (\%) are shown.}
\begin{tabular}{lccc}
\toprule
Method & 10\% Data & 50\% Data & 100\% Data \\
\midrule
LAPA  & 50.0 & 51.0 & 52.1 \\
ConLA & 58.3 & 60.4 & 64.6 \\
\bottomrule
\end{tabular}
\label{tab:data_scaling}
\end{table}

%% file: sec/5_Summary.tex
\section{Conclusion}
In this work, we propose ConLA, a simple yet effective method to extract high-quality latent actions from human demonstration videos for Vision-Language-Action models. By leveraging contrastive latent action learning with action-category and temporal priors to build action-centric representations, our approach mitigates shortcut learning and yields robust latent actions. Extensive experiments demonstrate that ConLA consistently outperforms previous methods, even when trained solely on human video data. These results demonstrate the strong potential of large-scale human video pre-training for VLA.

%% file: sec/6_Supplementary.tex
{\large{\noindent\textbf{Appendix}}}

\section{Implementation Details}
\subsection{Detailed Algorithm}

Algorithm~\ref{alg:latent action model} shows the detailed procedure of Contrastive Latent Action Learning. Algorithm~\ref{alg:pretain and finetune} provides the pseudocode for latent action policy pretraining and action finetuning. As shown in Algorithm~\ref{alg:latent action model}, prior to performing contrastive latent action learning, we first conduct a warm-up phase of 5,000 steps. During Warmup, we optimize model solely using the reconstruction loss. This is because at the beginning of training, the model has not yet learned a stable representation, and applying contrastive learning at this stage may lead to model collapse. By first optimizing the reconstruction loss, the model can acquire preliminary latent representations, which are then used to guide contrastive latent action learning and enhance the motion representations.

Algorithm~\ref{alg:pretain and finetune} demonstrates the pretraining and finetuning procedure of our policy. During the pretraining stage, we leverage the encoder trained in the first stage to infer latent action sequences from unlabeled videos, generating pseudo-labels. The policy is trained to predict the upcoming latent action sequence based on the current frame’s observation and instruction. During the finetuning stage, the policy is further trained on real robot trajectories, using the current observation and instruction, which enables alignment between latent actions and actual executed actions. Throughout the entire training process, the policy is based on the 7B Large World Model~\cite{liu2024world}. Both the pretraining and finetuning stages follow the LAPA~\cite{yelatent} framework; for more detailed implementation, please refer to LAPA~\cite{yelatent}.

Table~\ref{tab:hyperparameters} presents the training hyperparameter settings used in the first-stage contrastive latent action learning. We set the temperature coefficients for both action-centric and vision-centric contrastive learning to 0.07. To facilitate a fair comparison with our baseline LAPA~\cite{yelatent} and to highlight the effectiveness of our method, we keep our hyperparameters consistent with those used in LAPA~\cite{yelatent}. Since human-captured videos contain more motion-irrelevant noise, we follow LAPA’s~\cite{yelatent} design when choosing the frame interval: the frame interval for Something-SomethingV2 is set to 30, while that for BridgeV2 is set to 5.

\begin{algorithm}[thb]
\caption{ {\small Contrastive Latent Action Learning} }
\label{alg:latent action model}
\begin{algorithmic}[1] 
\footnotesize 
\State \textbf{Input:} $\mathcal{V}_{\text{unlabeled}}, Y_{\text{cls}}, {\text{Encoder:}}I_\phi, {\text{Decoder:}} F_\psi$
\State $\mathcal{V}_{\text{unlabeled}}$: unlabeled video $(O_t, I_t)$ pairs (observation, instruction)
\State $Y_{\text{cls}}$: Action class labels
\State $N_w$: number of Warmup update steps
\State $N_C$: number of ConLA update steps
\State \textbf{for} iter = 1 to $N_C$ \textbf{do}
\State \hspace{1em} Sample $(O_t,O_{t+k})$ and $(O_{t+k},O_t)$  from $\mathcal{V}_{\text{unlabeled}}$
\State \hspace{1em} $Z = I_\phi(\cdot \mid O_t, O_{t+k});[Z_{a^{\prime}}; Z_{v^{\prime}}] = \text{Split}(Z)$
\State \hspace{1em} $Z_a=\text{MLP}_{\text{action}}(Z_{a^{\prime}});Z_v=\text{MLP}_{\text{action}}(Z_{v^{\prime}})$
\State \hspace{1em} \textbf{if} iter$<N_w$ \textbf{then}
\State \hspace{2em} $\hat{O}_{t+k} = F_\psi(\cdot\mid  O_t,Z_a)$
\State \hspace{2em} $L_{\text{total}}=L_{\text{MSE}}(\phi,\psi)= \| \hat{O}_{t+k} - O_{t+k} \|^2$
\State \hspace{1em} \textbf{else}
\State \hspace{2em} $Z^{I} = I_\phi(\cdot \mid O_{t+k},O_t);[Z^{I}_{a^{\prime}}; Z^{I}_{v^{\prime}}] = \text{Split}(Z^{I})$
\State \hspace{2em} $Z^{I}_a=\text{MLP}_{\text{action}}(Z^{I}_{a^{\prime}});Z^{I}_v=\text{MLP}_{\text{visual}}(Z^{I}_{v^{\prime}})$
\State \hspace{2em} $\hat{O}_{t+k} = F_\psi(\cdot\mid  O_t,Z_a)$
\State \hspace{2em} $L_{\text{MSE}}(\phi,\psi)= \| \hat{O}_{t+k} - O_{t+k} \|^2$
\State \hspace{2em} $L_{\text{action}} = L_{\text{SupContrast}}(Z_a,Y_{\text{cls}})$ (Eq.~\ref{eq:action loss})
\State \hspace{2em} $L_{\text{visual}} = L_{\text{InfoNCE}}(Z_v,Z^I_{v})$ (Eq.~\ref{eq:visual loss})
\State \hspace{2em} $L_{\text{total}} = L_{\text{MSE}}
+ L_{\text{action}} + L_{\text{visual}}$
\State \hspace{1em} \textbf{end if}
\State \textbf{end for}
\end{algorithmic}
\end{algorithm}

\begin{algorithm}[thb]
\caption{ {\small Latent Action Pretraining \& Action Finetuning }}
\label{alg:pretain and finetune}
\begin{algorithmic}[1] 
\footnotesize 
\State \textbf{Input:} $\mathcal{V}_{\text{unlabeled}}, {\text{Encoder:}}I_\phi, \mathcal{D}_{\text{labeled}}, {\text{Latent Action Policy }} P_\theta$
\State $\mathcal{V}_{\text{unlabeled}}$: unlabeled video $(O_t, I_t)$ pairs (observation, instruction)
\State $\mathcal{D}_{\text{labeled}}$: real action trajectory $(O_t, I_t, A_t)$ pairs for fine-tuning
\State $N_P$: number of policy pretraining update steps
\State $N_F$: number of policy finetuning update steps
\State \textbf{Latent Action Pretraining}
\State \textbf{for} iter = 1 to $N_P$ \textbf{do}
\State \hspace{1em} Sample $(O_t,I_{t},Z^t_{a})$
from $\mathcal{V}_{\text{Pseudo}}$
\textbf{where} $Z^t_a = I_\phi(O_t, O_{t+k})$
\State \hspace{1em} $\hat{Z}^t_a = P_\theta(O_t,I_t)$
\State \hspace{1em} $L_{\text{MSE}}(\theta)= \| \hat{Z}^t_a - Z^t_a \|^2$
\State \textbf{end for}
\State \textbf{Action Finetuning}
\State \textbf{for} iter = 1 to $N_F$ \textbf{do}
\State \hspace{1em} Sample $(O_t,I_{t},A_t)$
from $\mathcal{D}_{\text{labeled}}$
\State \hspace{1em} $\hat{A}_t = P_\theta(O_t,I_t)$
\State \hspace{1em} $L_{\text{MSE}}(\theta)= \| \hat{A}_t - A_t \|^2$
\State \textbf{end for}
\end{algorithmic}
\end{algorithm}

\begin{table}[thb]
\centering
\caption{Latent action model Hyperparameters}
\label{tab:hyperparameters}
\begin{tabular}{l c} 
\toprule
\textbf{Hyperparameter} & \textbf{Value} \\
\midrule
Optimizer & AdamW \\
Learning Rate & 1e-4 \\
Batch Size  & 96 \\
Num Warmup updates &5000 \\
Num training updates &100000 \\
Embedding Dimension & 1024 \\
Quantization Dimension & 32 \\
Codebook Size & 8 \\
latent action Sequence Length & 4 \\
Contrastive Temperature ($\tau$) & 0.07 \\
Frame interval on SomethingV2  & 30\\
Frame interval on BridgeV2 & 5\\
\bottomrule
\end{tabular}
\end{table}

\subsection{Pre-training Dataset Processing}
We leverage natural language instructions as a bridge to extract structured action class labels, as instructions are highly correlated with executable action categories. Natural language conveys rich motion and spatial semantics, which can be distilled into explicit action category signals, providing clear supervision for latent action learning. This enables the automatic generation of action class labels from videos without ground-truth annotations, thereby supporting downstream contrastive latent action learning or policy learning. Our data preprocessing pipeline consists of the following stages:\\
(1) \textbf{Instruction normalization:} All instructions are converted to lowercase, and non-alphanumeric characters are removed. Sentences containing conjunctions (e.g., “and”) are filtered out, as such sentences typically describe multiple actions, which complicates the classification of atomic actions.\\
(2) \textbf{Action extraction:} Tokenization and part-of-speech tagging are performed using SpaCy (en\_core\_web\_lg). SpaCy is an efficient natural language processing library that supports tokenization, POS tagging, and dependency parsing. We use it to identify the main verb in each instruction as the core action information.\\
(3) \textbf{Spatial Direction Mapping:} Directional keywords (e.g., “top”, “left”, “in front of”) are mapped to a standardized set of direction categories using a manually constructed dictionary.\\
(4) \textbf{Label composition:} Each instruction is represented as a (verb, direction) pair, forming a discrete action label.\\
(5) \textbf{Data cleaning and category consolidation:} Instructions lacking valid verbs, containing ambiguous semantics, or having insufficient textual content are discarded. Classes with sample counts below a minimum threshold are merged into an "uncertain" category.

BridgeV2~\cite{walke2023bridgedata} does not provide action category labels. For this dataset, we apply the aforementioned preprocessing pipeline to categorize its natural language instructions into 80 discrete action classes. In contrast, the Something-SomethingV2~\cite{goyal2017something} dataset includes predefined action category labels for each video clip, with a total of 174 action classes. Despite the simplicity of this pipeline, the resulting pseudo action class labels are sufficiently stable and coherent to effectively support contrastive latent action learning. In future work, we will investigate automated approaches for extracting more fine-grained action category labels from both video and natural language instructions, with the goal of further improving the performance of contrastive latent action learning.

\section{Detailed Experimental}
\subsection{SimplerEnv}
\textbf{Experiment Setup.} SimplerEnv does not provide trajectories for fine-tuning, we follow the experimental setup of LAPA~\cite{yelatent}. In particular, once the base VLM is trained on BridgeV2~\cite{walke2023bridgedata}, we perform rollouts on the four SIMPLER~\cite{li2025evaluating} tasks and filter out 25 successful trajectories per task, resulting in 100 fine-tuning trajectories. Importantly, the trajectories used for fine-tuning differ from the evaluation setup in both object orientation and position. During evaluation, we conduct 24 rollouts for each task while randomizing the initial object locations. The average success rate is reported as the evaluation metric.\\

\noindent\textbf{Results.} We provide the evaluation results of baselines on the Simpler. We introduce a new baseline, UniVLA~\cite{bu2025univla}, which leverages Dinov2~\cite{oquab2024dinov2} features reconstructed from future frames to mitigate environmental noise and to construct task-centric representations that enhance latent action learning. To fairly compare the quality of latent actions learned in the first stage of latent action learning, the base model of UniVLA~\cite{bu2025univla} is aligned with ours ConLA and LAPA~\cite{yelatent}, all using the same Large World Model–7B~\cite{liu2024world}. The results report detailed performance across four tasks—stacking the green block onto the yellow block, placing the carrot on the plate, placing the spoon on the towel, and putting the eggplant into the basket—as well as their subtasks (grasping and moving). Tables~\ref{tab:simple_bridge_results} and ~\ref{tab:simple_something_results} present results on BridgeV2~\cite{walke2023bridgedata} and human manipulation videos, respectively. UniVLA~\cite{bu2025univla} achieves performance comparable to LAPA~\cite{yelatent} on BridgeV2~\cite{walke2023bridgedata}, while obtaining a substantially larger improvement on human videos. This indicates that UniVLA~\cite{bu2025univla} is effective at extracting higher-quality latent actions under complex environmental variations present in human demonstrations. However, due to the lack of inductive biases, UniVLA~\cite{bu2025univla} remains susceptible to interference from irrelevant visual information, which restricts its ability to further improve performance.

\subsection{Real-World Robots}
\begin{figure}[H]
    \centering
    \includegraphics[width=1.0\columnwidth]{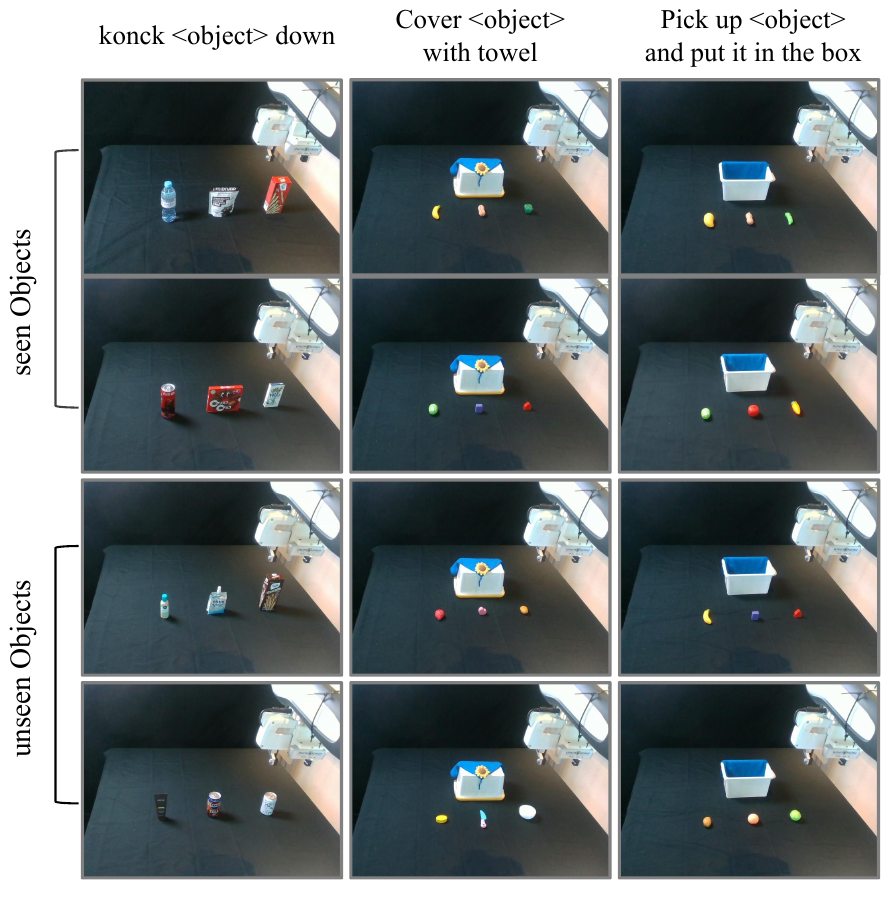}
    \caption{Real-world Manipulation Examples}
    \label{fig:real-world}
\end{figure}

\textbf{Experiment Setup.} To enable a direct comparison with our baselines, we follow the real-world experimental setup of LAPA~\cite{yelatent}. Figure~\ref{fig:real-world} illustrates sample executions of each real-world tabletop manipulation task. For teleoperation data collection, we use GELLO to collect 150 trajectories per task. Each scene contains three objects, and the model must determine which object to interact with based on the task instruction. For each task, we evaluate three distinct capabilities:(1) Generalization to unseen combinations of previously seen objects during fine-tuning. (2) Generalization to completely unseen objects during fine-tuning, which may or may not have been observed during pretraining. (3) Generalization to unseen instructions requiring semantic reasoning. For each evaluation criterion, we perform 6 rollouts, resulting in 18 rollouts per task category. Since there are three tasks, each model is evaluated with a total of 54 real-world rollouts. For fair comparison, we use an identical image resolution across all models and keep the initial positions of all objects fixed. For evaluation metrics, we adopt the same partial success criteria as LAPA~\cite{yelatent} to enable fine-grained assessment. The detailed scoring scheme is provided below.

\textbf{Knock down the \textless object\textgreater.}
The robot receives 0.5 score for reaching the correct object and 1 score for successfully knocking it down.

\textbf{Cover the \textless object\textgreater with a towel.} The robot receives 0.33 score for successfully picking up the towel, 0.66 score for reaching the correct object and partially covering it, and 1 score for fully covering the target object.

\textbf{Pick up the \textless object\textgreater and put it in the box.} The robot receives 0.25 score for reaching the correct object, 0.5 score for successfully grasping it, 0.75 score for grasping and moving it toward the box without successfully placing it, and 1 score for correctly placing the object into the box.\\

\noindent\textbf{Results.} Tables~\ref{tab:knocking},~\ref{tab:Covering}, and~\ref{tab:pick place} provide the full list of objects used in the evaluation rollouts for the Knocking, Covering, and Pick \& Place tasks, respectively, along with the corresponding partial success scores. Table~\ref{tab:total_success_rates} reports the overall average success rate. From the experimental results, we observe clear performance gains over the Scratch baseline for both robot-video and human-video pretraining, demonstrating the effectiveness of pretraining. Moreover, in the unseen object setting across all three tasks—knock, cover, and pick and place—human-video pretraining consistently outperforms BridgeV2~\cite{walke2023bridgedata} pretraining for both LAPA~\cite{yelatent} and ConLA. A explanation is that some of the unseen objects may appear in the human-video pretraining corpus, enabling stronger generalization and highlighting the potential of human-video pretraining. Additionally, we report the strict success rate in the tables, where our model achieves higher strict success than LAPA~\cite{yelatent}. We attribute this improvement to our ability to extract higher-quality latent actions, which more effectively transfer motion priors into the downstream policy.

\section{More Visualization}
\begin{figure}[H]
   \includegraphics[width=1.0\linewidth]{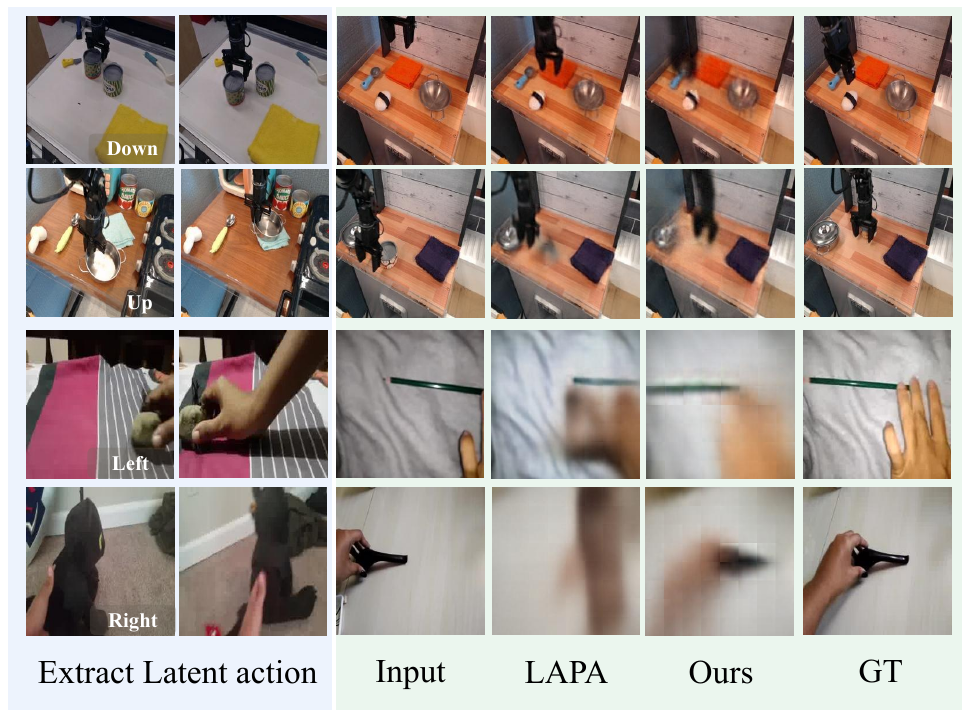}
\caption{Latent action consistency visualization analysis}
\label{fig:latent action vis}
\end{figure}

Figure~\ref{fig:latent action vis} presents additional visualizations for latent action consistency. From these results, we clearly observe that LAPA~\cite{yelatent} exhibits significant latent action inconsistency, particularly when trained on human video data where shortcut learning is more likely to occur. 

In our analysis, we extract the latent actions corresponding to left and right motion from two image pairs, and then apply each extracted latent action to a new starting frame to reconstruct the expected motion outcome. For LAPA~\cite{yelatent}, both the “left” and “right” reconstructions inadvertently reproduce visual content from the frames used to extract the latent actions, indicating that the extracted representation encodes visual appearance rather than motion, which is a direct symptom of shortcut learning. In contrast, our method successfully extracts motion-centric latent actions and reconstructs the intended motion outcomes without leaking appearance information.

For the robot-video setting—which contains more controlled scenes and less visual noise—shortcut learning is less problematic. However, even in this cleaner setting, LAPA~\cite{yelatent} still shows latent action inconsistencies. For example, in the first row, when extracting a horizontal-down motion, LAPA~\cite{yelatent} incorrectly captures a vertical-down motion, whereas our method correctly captures the horizontal-down direction. In the second row, LAPA~\cite{yelatent} reconstructs an upper-left motion, while our method more accurately extracts the upward motion, demonstrating better alignment between the intended and extracted latent actions.

\clearpage
\begin{table*}[thb]
\centering
\caption{\textbf{SimplerEnv results of Bridgev2 Pretraining.} We pretrain baselines on BridgeV2 video dataset. Here, We reproduce the UniVLA* results using the Large World Model 7B. The table reports Success, Grasping, and Moving rates(\%). The four evaluated tasks are: stack green to yellow block, put carrot on plate, put spoon on towel, and put eggplant in basket.}
\label{tab:simple_bridge_results}
\renewcommand{\arraystretch}{1.0}

\begin{tabular}{lccccccc}
\toprule
\textbf{Success Rate} & Scratch & UNIPI & VPT & LAPA & UniVLA* & ConLA & ActionVLA \\
\midrule
StackG2Y        & 29.2 & 2.7 & 45.8 & 54.2 & 41.7 & 62.5 & \textbf{75.0} \\
Carrot2Plate     & 29.2 & 2.7 & 37.5 & 45.8 & 45.8 & 45.8 & \textbf{58.0} \\
Spoon2Towel      & 50.0 & 0.0 & 70.8 & 70.8 & 75.0 & \textbf{75.0} & 70.8 \\
Eggplant2Bask    & 29.2 & 0.0 & 50.0 & 58.3 & \textbf{62.5} & 58.3 & 50.0 \\
\midrule
AVG              & 34.4 & 1.3 & 51.0 & 57.3 & 56.2 & 60.4 &\textbf{63.5} \\

\midrule
\textbf{Grasping Rate} & Scratch & UNIPI & VPT & LAPA & UniVLA* & ConLA & ActionVLA \\
\midrule
Grasp Green Block & 66.6 & 20.8 & 62.5 & 62.5 & 58.3 & 62.5 &\textbf{87.5} \\
Grasp Carrot      & 45.8 & 33.2 & 54.1 & 58.3 & 46.8 & 45.8 & \textbf{75.0} \\
Grasp Spoon       & 70.8 & 22.2 & 79.2 & 83.3 & 75.0 & 75.0 & \textbf{83.3} \\
Grasp Eggplant    & 62.5 & 16.0 & 70.8 & \textbf{83.3} & 79.2 & 75.0 & 75.0 \\
\midrule
AVG               & 61.4 & 23.1 & 66.7 & 71.9& 64.8 & 64.6 & \textbf{80.2} \\

\midrule
\textbf{Moving Rate} & Scratch & UNIPI & VPT & LAPA & UniVLA* & ConLA & ActionVLA\\
\midrule
Move Green Block  & 58.3 & 29.1 & 58.3 & 66.6 & 58.3 & 62.5 & \textbf{91.6} \\
Move Carrot       & 45.8 & 48.6 & 66.6 & 75.0 & 50.0 & 54.2 & \textbf{91.6} \\
Move Spoon        & 70.8 & 34.6 & 79.2 & \textbf{83.3}  & 75.0 & 75.0 & 79.2 \\
Move Eggplant     & 87.5 & 58.0 & 70.8 & 87.5  & 79.2 & 83.3 & \textbf{91.6} \\
\midrule
AVG               & 65.6 & 42.6 & 68.7 & 77.1 & 65.6 & 68.8 &\textbf{88.5} \\
\bottomrule
\end{tabular}
\end{table*}

\begin{table*}[thb]
\centering
\caption{\textbf{SimplerEnv results of Human Manipulation Video Pretraining.} We pretrain baselines on the Something-SomethingV2 video dataset. Here,We reproduce the UniVLA* results using the Large World Model 7B. The table reports Success, Grasping, and Moving rates(\%). The four evaluated tasks are: stack green to yellow block, put carrot on plate, put spoon on towel, and put eggplant in basket.}
\label{tab:simple_something_results}
\renewcommand{\arraystretch}{1.0}

\begin{tabular}{lcccccc}
\toprule
\textbf{Success Rate}& Scratch & UNIPI & VPT & LAPA & UniVLA* & ConLA\\
\midrule
StackG2Y        &29.2 & 0.0 & 50.0 & 50.0 & 62.5 &  \textbf{62.5}\\
Carrot2Plate     &29.2 & 1.3 & 29.1 & 50.0 & 37.5 &  \textbf{50.0}\\
Spoon2Towel      &50.0 & 1.3 & 37.5 & 50.0 & 70.8 &  \textbf{79.2}\\
Eggplant2Bask    &29.2 & 0.0 & 66.6 & 58.3 & 50.0 &  \textbf{66.6}\\
\midrule
AVG              &34.4  & 0.7 & 45.8 & 52.1 & 55.2 & \textbf{64.6}\\

\midrule
\textbf{Grasping Rate} & Scratch & UNIPI & VPT & LAPA & UniVLA* & ConLA\\
\midrule
Grasp Green Block & 66.6 & 2.7  & 66.6 & 58.3 & \textbf{66.7} & 62.5\\
Grasp Carrot      & 45.8 & 31.7 & 45.8 & \textbf{62.5} & 45.8 & 45.8\\
Grasp Spoon       & 70.8 & 21.7 & 70.8 & 75.0  & 75.0 & \textbf{87.5}\\
Grasp Eggplant    & 62.5 & 6.8  & \textbf{91.6} & 70.8 & 62.5 & 75.0\\
\midrule
AVG               & 61.4 & 15.7 & \textbf{68.7} & 66.7 & 62.5 & 67.7\\

\midrule
\textbf{Moving Rate} & Scratch & UNIPI & VPT & LAPA &UniVLA* & ConLA \\
\midrule
Move Green Block  & 58.3 & 2.7  & 62.5  & 62.5 & 62.5  & \textbf{62.5}  \\
Move Carrot       & 45.8 & 37.5 & 58.3  & \textbf{70.8} & 54.2  & 58.3  \\
Move Spoon        & 70.8 & 18.1 & 54.1  & 75.0 & 83.3  & \textbf{87.5}  \\
Move Eggplant     & 87.5 & 50.3 & 91.6  & \textbf{93.3} & 75.0 & 79.2  \\
\midrule
AVG               & 65.6 & 27.1 & 66.6 &\textbf{ 72.9} & 68.8  &  71.9   \\
\bottomrule
\end{tabular}
\end{table*}

\clearpage
\begin{table}[thb]
\centering
\captionsetup{
  singlelinecheck=false,
  font=bf,
  labelfont=bf,
}
\caption{Knocking Task Results}
\label{tab:knocking}
\begin{tabular}{l c c c c c}
\toprule
 & Scratch & LAPA (Bridge) & ConLA (Bridge) & LAPA (Sthv2) & ConLA (Sthv2) \\
\midrule
\multicolumn{6}{c}{\textbf{Seen Objects, Unseen Object Combinations}} \\
\midrule
bottle    & 0.5 & 0   & 0   & 0    & 1\\
chocolate & 0   & 0   & 1   & 0.5  & 1\\
crisp     & 0   & 0.5 & 0.5 & 0    & 0\\
cocacola  & 0.5 & 0   & 0.5 & 0.5  & 0\\
pie       & 0   & 0   & 0   & 0.5  & 0.5\\
pocky     & 0.5 & 1   & 1   & 1    & 1\\
\midrule
SUM & 1.5 & 1.5 & 3 & 2.5 & 3.5\\
\midrule

\multicolumn{6}{c}{\textbf{Unseen Objects}} \\
\midrule
pepsi           & 0   & 0   & 1   & 1   & 1\\
conditioner     & 0   & 0   & 0   & 0   & 0\\
CALPIS          & 0   & 0   & 0   & 0   & 0\\
grey-chocolate  & 0   & 1   & 0   & 0   & 0.5\\
milk-tea        & 0 & 0   & 0.5 & 0   & 1\\
shampoo         & 0   & 0   & 0   & 0   & 0\\
\midrule
SUM & 0 & 1 & 1.5 & 1 & 2.5\\
\midrule

\multicolumn{6}{c}{\textbf{Seen Objects, Unseen Instructions}} \\
\midrule
pillared object     & 0   & 0 & 1   & 0   & 0 \\
red-packed food     & 0   & 0 & 0   & 0.5 & 1\\
white-bagged snacks & 0   & 1 & 1   & 0.5 & 0\\
carbonated drinks   & 0.5 & 1 & 0.5 & 1   & 1\\
cookie box          & 0.5 & 1 & 0   & 1   & 1\\
rectangle object    & 0   & 0 & 0   & 0   & 0.5\\
\midrule
SUM & 1 & 3 & 2.5 & 3 & 3.5\\
\midrule
Success Rate (Strict) & 0\% & 33.33\% & 27.78\% & 27.78 \% & 44.44\%\\
Success Rate & 13.89\% & 30.56\% & 38.89\% & 36.11\% & 52.78\%\\
Reaching Success Rate & 27.78\% & 33.33\% & 50\% & 50\% & 61.11\%\\
\bottomrule
\end{tabular}
\end{table}

\clearpage

\begin{table}[thb]
\centering
\captionsetup{
  singlelinecheck=false,
  font=bf,
  labelfont=bf,
}
\caption{Covering Task Results}
\label{tab:Covering}
\begin{tabular}{l c c c c c}
\toprule
 & Scratch & LAPA (Bridge) & ConLA (Bridge) & LAPA (Sthv2) & ConLA (Sthv2) \\
\midrule
\multicolumn{6}{c}{\textbf{Seen Objects, Unseen Object Combinations}} \\
\midrule
banana        & 0.33  & 0.33   & 0.66   & 0.33    & 0.66\\
peanut        & 0     & 0.33   & 0.33   & 0.33    & 0.33\\
pepper        & 0.33  & 0.33   & 0.33   & 0.33    & 0.66\\
cabbage       & 0.33  & 0.33   & 0.66   & 0.66    & 1\\
purple-block  & 0     & 0.66   & 0.33   & 0.33    & 0.33\\
red-block     & 0.33  & 1      & 1      & 0       & 0.66\\
\midrule
SUM & 1.32 & 1.98 & 3.31 & 1.98 & 3.64\\
\midrule

\multicolumn{6}{c}{\textbf{Unseen Objects}} \\
\midrule
strawberry           & 0.66   & 0.66   & 0.33   & 0.33   & 1\\
potato               & 0.33   & 0      & 0.33   & 0.33   & 0.33\\
heart-shaped block   & 0.33   & 0.33   & 0.33   & 0.66   & 0.33\\
oval block           & 0      & 0.33   & 0.66   & 1      & 1\\
knife                & 0.33   & 0.66   & 0      & 1      & 1\\
bowl                 & 0      & 0      & 0.66   & 0.33   & 0.33\\
\midrule
SUM & 1.65 & 1.98 & 2.31 & 2.65 & 3.99 \\
\midrule

\multicolumn{6}{c}{\textbf{Seen Objects, Unseen Instructions}} \\
\midrule
yellow fruit        & 0.33   & 0    & 0.33   & 0.33   & 0.66 \\
green vegetable     & 0.33   & 0.33 & 0.66   & 0.33   & 0.66\\
nut                 & 0      & 0.33 & 0.33   & 0.66   & 0.33\\
spicy vegetable     & 0      & 0    & 0      & 0      & 0\\
rectangle object    & 0.33   & 0.66 & 0.33   & 0.33   & 0.33\\
polygonal block     & 0.33   & 0.33 & 0.66   & 0.66   & 0.66\\
\midrule
SUM & 1.32 & 1.65 & 2.31 & 2.31 & 2.64 \\
\midrule
Success Rate (Strict) & 0\% & 5.5\% & 5.5\% & 11.11 \% & 22.22\%\\
Success Rate & 23.83\% & 36.72\% & 44.06\% & 38.56\% & 57.06\%\\
Reaching Success Rate & 5.56\% & 27.78\% & 38.89\% & 33.33\% & 50\%\\
\bottomrule
\end{tabular}
\end{table}

\clearpage

\begin{table}[thb]
\centering
\captionsetup{
  singlelinecheck=false,
  font=bf,
  labelfont=bf,
}
\caption{Pick \& Place Box Task Results}
\label{tab:pick place}
\small
\begin{tabular}{l c c c c c c c}
\toprule
 & Scratch & LAPA (Bridge) & ConLA (Bridge) & LAPA (Sthv2) & ConLA (Sthv2) \\
\midrule
\multicolumn{6}{c}{\textbf{Seen Objects, Unseen Object Combinations}} \\
\midrule
apple        & 0.25  & 0.25   & 0.25   & 0.25    & 0.5\\
bean         & 0     & 1      & 0.75   & 0.75    & 1\\
cabbage      & 0     & 0      & 0      & 0       & 0.75\\
carrot       & 0     & 0.75   & 1      & 1       & 1\\
mango        & 0.25  & 0      & 0      & 0       & 0.25\\
peanut       & 0     & 0      & 0      & 0       & 0\\
\midrule
SUM & 0.5 & 2 & 2 & 2 & 3.5\\
\midrule

\multicolumn{6}{c}{\textbf{Unseen Objects}} \\
\midrule
tomato           & 0      & 0.25   & 0.25   & 0.5    & 1\\
peach            & 0      & 0      & 0      & 0      & 0\\
avocado          & 0      & 0.25   & 0.25   & 0.25   & 0.25\\
banana           & 0.25   & 0      & 0      & 0.25   & 0.5\\
purple-block     & 0      & 0.25   & 0      & 0      & 0\\
red-block        & 0      & 0.25   & 0.25   & 0      & 0.25\\
\midrule
SUM & 0.25 & 1 & 0.75 & 1 & 2 \\
\midrule

\multicolumn{6}{c}{\textbf{Seen Objects, Unseen Instructions}} \\
\midrule
an object that is red          & 0.55 & 0    & 0     & 0     & 0 \\
an object that is green        & 0    & 0.25 & 0.5   & 0     & 0.25\\
an object that is a vegetable  & 0    & 1    & 1     & 0.5   & 0.25\\
an object that is orange       & 0.25 & 0.5  & 0.25  & 0.5   & 0.25\\
an object that is yellow       & 0    & 0    & 0.25  & 0     & 0\\
nut                            & 0    & 0    & 0     & 0     & 0\\
\midrule
SUM & 0.75 & 1.75 & 2 & 1 & 0.75 \\
\midrule
Success Rate (Strict) & 0\% & 11.11\% & 11.11\% & 5.6 \% & 16.67\%\\
Success Rate & 8.33\% & 26.39\% & 26.39\% & 22.22\% & 34.72\%\\
Reaching Success Rate & 27.78\% & 55.56\% & 55.56\% & 44.44\% & 66.67\%\\
\bottomrule
\end{tabular}
\end{table}

\begin{table}[htb]
\centering
\captionsetup{
  singlelinecheck=false,
  font=bf,
  labelfont=bf,
  textfont=normalsize
}
\caption{Summary of Total Success Rates (\%)}
\label{tab:total_success_rates}
\begin{tabular}{l c c c c c}
\toprule
 & Scratch & LAPA(Bridge) & ConLA (Bridge) & LAPA (Sthv2) & ConLA (Sthv2) \\
\midrule
Total Success Rate & 15.35\% & 31.22\% & 36.45\% & 32.30\% & \textbf{48.18\%} \\
Total Success Rate (Strict) & 0\% & 14.80\% & 14.80\% & 14.83\% & \textbf{27.78\%} \\
\bottomrule
\end{tabular}
\end{table}